\documentclass{article}
\usepackage{ijcai26}

\usepackage{times}
\usepackage{soul}
\usepackage{url}
\usepackage[hidelinks]{hyperref}
\usepackage[utf8]{inputenc}
\usepackage[small]{caption}
\usepackage{graphicx}
\usepackage{amsmath}
\usepackage{amsthm}
\usepackage{booktabs}
\usepackage{algorithm}
\usepackage{algorithmic}
\usepackage[switch]{lineno}
\usepackage{subcaption}

\title{ExPosST: Explicit Positioning with Adaptive Masking for LLM-Based Simultaneous Machine Translation}

\author{
Yuzhe Shang$^1$\thanks{This work was partially done when Yuzhe Shang was an intern at MiLM Plus team.}
\and
Pengzhi Gao$^2$\and
Yazheng Yang$^3$\and
Jiayao Ma$^1$\and\\
Wei Liu$^2$\and
Jian Luan$^2$\and
Jinsong Su$^1$
\\
\affiliations
$^1$School of Informatics, Xiamen University, China\\
$^2$MiLM Plus, Xiaomi Inc., Beijing, China\\
$^3$The University of Hong Kong, Hong Kong, China\\
\emails
\{shangyuzhe, majiayao\}@stu.xmu.edu.cn, 
jssu@xmu.edu.cn, 
\{gaopengzhi,liuwei40,luanjian\}@xiaomi.com, 
oldbirdaz@gmail.com
}

\begin{document}

\maketitle

\begin{abstract}
Large language models (LLMs) have recently demonstrated promising performance in simultaneous machine translation (SimulMT). However, applying decoder-only LLMs to SimulMT introduces a positional mismatch, which leads to a dilemma between decoding efficiency and positional consistency. Existing approaches often rely on specific positional encodings or carefully designed prompting schemes, and thus fail to simultaneously achieve inference efficiency, positional consistency, and broad model compatibility. In this work, we propose ExPosST, a general framework that resolves this dilemma through explicit position allocation. ExPosST reserves fixed positional slots for incoming source tokens, enabling efficient decoding with KV cache across different positional encoding methods. To further bridge the gap between fine-tuning and inference, we introduce a policy-consistent fine-tuning strategy that aligns training with inference-time decoding behavior. Experiments across multiple language pairs demonstrate that ExPosST effectively supports simultaneous translation under diverse policies\footnote{Code will be made publicly available upon the publication of this paper.}.
\end{abstract}

\section{Introduction}
\label{sec:intro}
\begin{figure}[ht!]
    \centering
    \includegraphics[width=1\linewidth]{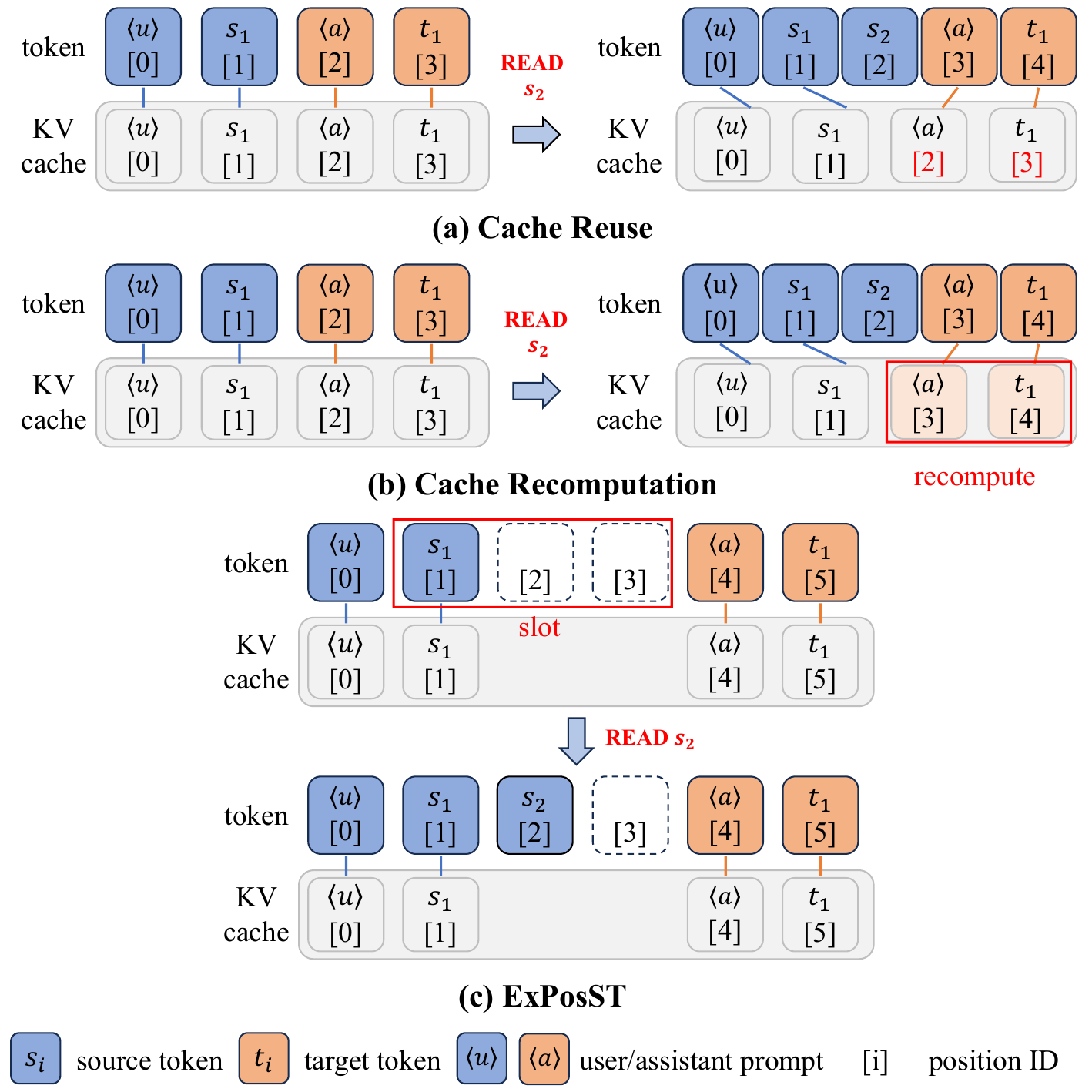}
    \caption{Comparison of different strategies for LLM-based SimulMT. $\langle u \rangle$ and $\langle a \rangle$ denote the user and assistant role prompts in the LLM conversational template. (a) \textbf{Cache Reuse}: Inserting a new source token ($s_3$) shifts the positional indices of subsequent tokens (e.g., $t_1$ shifts from 3 to 4), causing a mismatch with the existing KV cache. (b) \textbf{Cache Recomputation}: Positional consistency is restored by re-encoding the shifted tokens, but this incurs prohibitive computational latency. (c) \textbf{ExPosST (Ours)}: By explicitly pre-allocating positional slots for potential source tokens, the positional indices of the target sequence remain invariant during READ/WRITE cycles, enabling efficient KV cache reuse without positional misalignment.}
    \label{fig:position}
\end{figure}

Simultaneous Machine Translation (SimulMT) is a prominent sub-field in Neural Machine Translation (NMT), where target language output is generated in real time as parts of the source sentence arrive. By trading off translation quality against latency, SimulMT plays a critical role in real-world applications such as international conferences, live broadcasts, and academic lectures, and has therefore attracted sustained research interest \cite{ma-etal-2019-stacl,DBLP:conf/iclr/ZhangF23,agostinelli-etal-2024-simul}. Recently, large language models (LLMs) have demonstrated strong performance in offline NMT~\cite{alves-etal-2023-steering,DBLP:conf/iclr/Xu0SA24,zhu-etal-2024-fine}, motivating a growing body of work that explores their potential for SimulMT~\cite{agostinelli-etal-2024-simul,koshkin-etal-2024-transllama,xu-etal-2025-seqpo}. Despite their promise, directly applying decoder-only LLMs to SimulMT remains non-trivial and introduces new challenges that do not arise in conventional offline translation.

A central challenge lies in a fundamental \emph{positional dilemma} inherent to SimulMT with decoder-only LLMs~\cite{raffel-etal-2024-simultaneous}. While modern decoder-only LLMs leverage Key-Value (KV) caching to achieve efficient auto-regressive decoding, the streaming nature of SimulMT inherently disrupts this mechanism. In SimulMT, however, newly arriving source tokens must be inserted into the existing context, which shifts the positional indices of subsequent target tokens. As illustrated in Figure~\ref{fig:position}(a), this positional shift invalidates the cached KV states, leading to a mismatch between the positional information embedded in the KV cache and the updated token indices in the shifted sequence. While recomputing the KV cache can restore positional consistency (Figure~\ref{fig:position}(b)), it incurs prohibitive computational overhead and negates the efficiency benefits of caching. This creates an direct dilemma between inference efficiency and positional consistency, posing a major obstacle to practical LLM-based SimulMT.

Existing approaches attempt to mitigate this dilemma through two main directions, but none resolve it in a fully general manner. One line of work modifies internal model mechanisms, such as positional encodings and attention masks. For example, SimulMask~\cite{raffel-etal-2024-simultaneous} aligns fine-tuning and inference by restricting target-to-source attention and adopting a modified ALiBi positional encoding~\cite{press2022trainshorttestlong}. While effective under specific settings, such methods are tightly coupled with particular positional encodings and do not readily extend to mainstream RoPE-based LLMs such as Llama~\cite{dubey2024llama} and Qwen~\cite{yang2024qwen2} series. Group Position Encoding (GPE)~\cite{tong-etal-2025-llm} avoids cache recomputation by separating positional spaces for source and target tokens, but introduces overlapping or non-integer positions that are incompatible with standard LLM pretraining. Another line of work reformulates SimulMT as a conversation, appending source and target tokens sequentially and separating them using user--assistant roles~\cite{wang-etal-2024-simultaneous,fu2025llmsachievehighqualitysimultaneous}. Although this strategy enables efficient cache reuse, frequent role switching may distract the model from the source context and degrade translation fluency. As a result, achieving efficiency, positional consistency, and broad model compatibility simultaneously remains an open challenge.

To address these limitations, we propose \textbf{ExPosST} (\textbf{Ex}plicit \textbf{Pos}ition Allocation for \textbf{S}imultaneous \textbf{T}ranslation), a simple and general framework for LLM-based SimulMT. The key idea of ExPosST is to ensure the invariance of target token positions regardless of the incremental growth of source tokens. Concretely, ExPosST adopts a \emph{pre-allocated positions inference} strategy that pre-allocates fixed positional slots for potential source tokens, while generating target tokens from a subsequent invariant position range. As shown in Figure~\ref{fig:position}(c), this design ensures that target positional indices remain unchanged throughout READ/WRITE cycles, enabling direct KV cache reuse without recomputation while preserving strict positional consistency. In addition, we introduce a \emph{policy-consistent fine-tuning} strategy that aligns the training data format and attention masks with the slot-based inference behavior, eliminating the mismatch between fine-tuning and deployment. Extensive experiments across multiple language pairs, decoding policies, and mainstream RoPE-based LLMs---including \texttt{Llama-3.1-8B-Instruct} and \texttt{gemma-2-2b-it}---demonstrate that ExPosST achieves state-of-the-art translation quality at comparable latency, while exhibiting strong generality across models and positional encoding schemes.

\begin{figure*}[!t]
    \centering
    \includegraphics[width=1\linewidth]{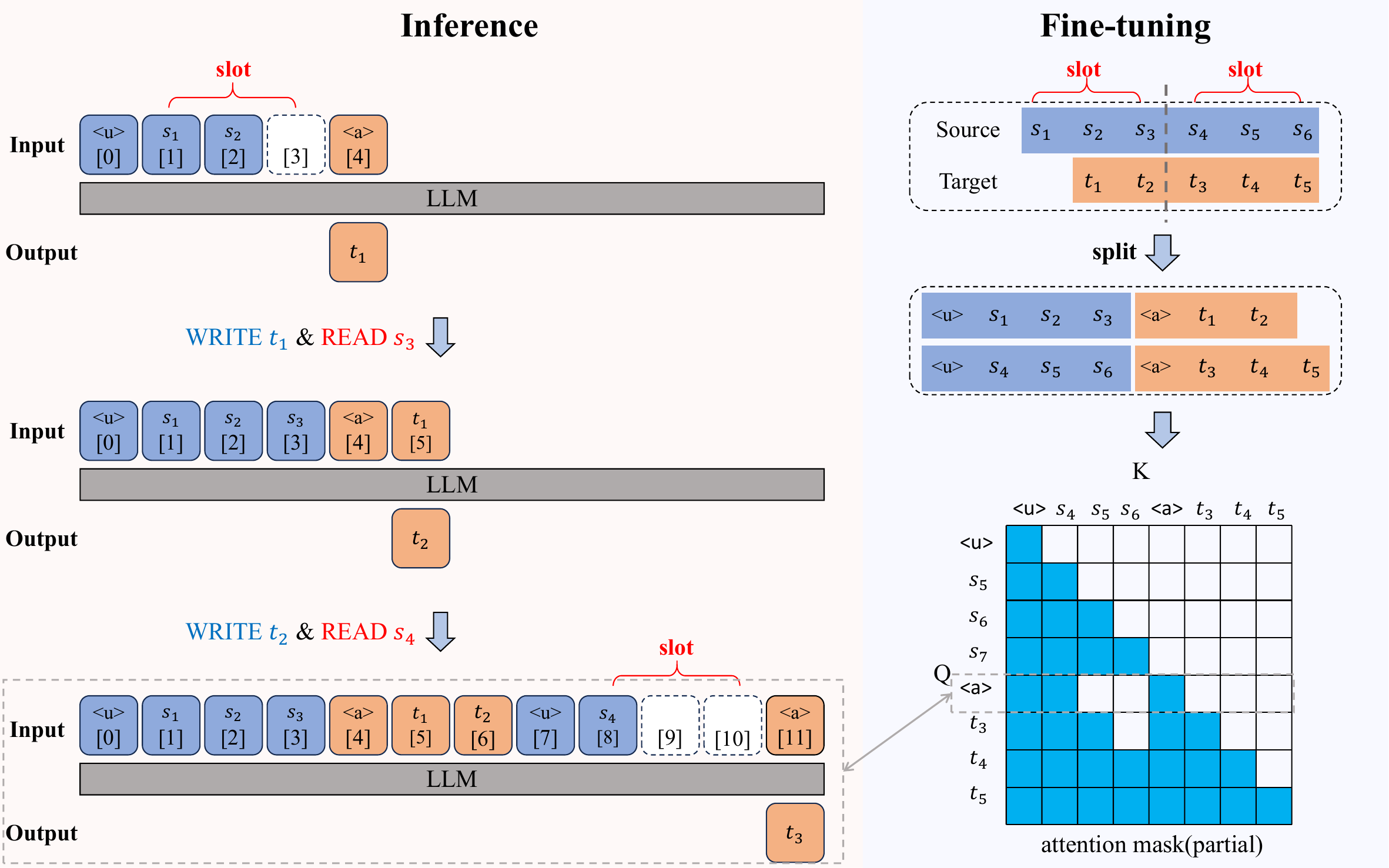}
    \caption{Overview of the ExPosST framework. \textbf{Left:} The \textit{Inference with Pre-allocated Positions} strategy. We adopt a wait-2 policy and set the pre-allocated slot length $L_{slot}=3$. When $s_3$ is read, it fills the reserved slot without shifting the positions of generated target tokens (e.g., $t_1$). Upon reading $s_4$, a new source slot is allocated immediately after the current output to maintain positional consistency. \textbf{Right:} The \textit{Policy-Consistent Fine-tuning} strategy. The source sentence is segmented into parts to match the inference slot. A policy-consistent attention mask (bottom right) is applied to ensure the visibility of source tokens aligns with the specific simultaneous policy.}
    \label{fig:main}
\end{figure*}

\section{Methodology}
In this section, we provide a detailed description of the ExPosST framework. We first provide a formal problem formulation and discuss the technical background of LLM-based SimulMT in Section~\ref{sec:background}, which clarifies the challenges of maintaining positional consistency. To address these challenges, we propose a novel architecture whose overall workflow is illustrated in Figure~\ref{fig:main}. The framework consists of two synergistic components: a \textit{Pre-allocated Positions Inference} strategy that enables zero-recomputation inference through position pre-allocation, and a \textit{Policy-Consistent Fine-tuning} strategy that strictly aligns training supervision with streaming behaviors.

\subsection{Task Formulation}
\label{sec:background}

SimulMT aims to generate a target sequence $T = (t_1, \dots, t_N)$ incrementally while reading a streaming source sequence $S = (s_1, \dots, s_M)$. Unlike offline translation, SimulMT requires the model to generate each target token $t_j$ based only on a partial source prefix available at each decoding step, formulated as $P(t_j | t_{<j}, s_{\le g(j)})$. 
Here $s_{\le g(j)}$ denotes the source prefix available at decoding step $j$, and $g(j)$ is a monotonic non-decreasing function determined by a read-write policy. A representative policy is \textit{wait-k} \cite{ma-etal-2019-stacl}, which first reads $k$ source words and then alternates between writing and reading. Another common policy is read-n \& incremental decoding policy \cite{wang-etal-2024-simultaneous}, where the model reads $n$ source tokens at each READ step before generating the next target segment.

Recently, the integration of LLMs has emerged as a prominent research direction within SimulMT. Typically, most LLMs adopt a decoder-only architecture, and utilize a KV cache to store the key ($K$) and value ($V$) matrices of previously processed tokens. This caching mechanism reduces the computational complexity of auto-regressive decoding from quadratic to linear. 
To encode positional information, mainstream LLMs employ Rotary Positional Embedding (RoPE)~\cite{SU2024127063}, applying a rotation transformation to the query ($Q$) and key ($K$) vectors based on their assigned positional index. 
Crucially, key vectors are cached after the RoPE transformation is applied. As a result, the positional information stored in the KV cache becomes fixed to their original positional indices. 
Consequently, any incremental update to the source prefix during inference shifts the relative positions of subsequent tokens, leading to a mismatch between the cached states and the updated sequence.

\subsection{Pre-allocated Positions Inference}
\label{sec:inference}

The Pre-allocated Positions Inference strategy is designed to resolve the dilemma between inference efficiency and positional consistency. The core idea is to pre-assign a fixed, reserved block of positional indices, termed a \textit{position slot} $S^{(i)}$, for incoming source tokens. By doing so, the positional indices assigned to the target sequence remain invariant throughout READ and WRITE cycles. This stability allows the KV cache for target tokens to be reused directly without any recomputation, thereby eliminating the computational overhead associated with positional misalignment.

In practice, we construct the inference prompt using a standard conversational format, where the source and target tokens are delimited by special tokens such as \texttt{$\langle user \rangle$} and \texttt{$\langle assistant\rangle$}. To prevent dynamically arriving source tokens from shifting the positional indices of already-generated target tokens, we explicitly allocate a contiguous range of positional indices of length $L_{slot}$ for the current allocation phase $i$. Even if the actual source tokens do not yet fill this entire slot, the target generation commences at the first position immediately after the reserved block.

Formally, the starting positional index of the target segment $pos(t_{start}^{(i)})$ is determined by the starting positional index of the current source slot $pos(s_{start}^{(i)})$:
\begin{equation}
    pos(t_{start}^{(i)}) = pos(s_{start}^{(i)}) + L_{slot}
    \label{eq:tgtpos}
\end{equation}

During inference, new source tokens are inserted within the current slot $S^{(i)}$, while newly generated target tokens are sequentially appended after the slot. As illustrated in the left part of Figure \ref{fig:main}, when the new token $s_3$ is encoded, the positional indices of the previously generated target token remain unchanged, allowing the reuse of zero-recomputation KV cache for the target segment.

If the length of the received source tokens exceeds the capacity of the current slot(s) (e.g., when reading $s_4$ in Figure ~\ref{fig:main}), we allocate a new slot $S^{(i+1)}$ of size $L_{slot}$ immediately after the current output. Consequently, the starting position of subsequent outputs is updated according to Equation~\ref{eq:tgtpos} with the index $i+1$. This dynamic slot allocation ensures that the positional indices of all previously generated target tokens remain invariant, effectively preventing the positions of output tokens from shifting as input tokens increase. Therefore, the KV cache never requires recomputation, maintaining high inference efficiency while preserving strict positional consistency.
\subsection{Policy-Consistent Fine-tuning}
\label{sec:finetuning}

To bridge the gap between conventional offline training and our streaming inference framework, we introduce a Policy-Consistent Fine-tuning strategy. This approach ensures that the model is optimized under constraints that precisely mirror the positional and visibility conditions encountered during real-time inference. 
Concretely, we conduct the following two adaptations:

First, we restructure the training data to reflect the pre-allocated position slot scheme used during inference. 
As illustrated in the right part of Figure \ref{fig:main}, the source sentence is segmented by slots~(e.g., $\{s_1, s_2, s_3\}$ and $\{s_4, s_5, s_6\}$) that correspond to the maximum capacity of a single position slot $L_{slot}$. This segmentation aligns the positional indices of source tokens in the training samples with the fixed slot-based structure employed at inference time. 
As a result, the model learns to generate target tokens starting from the predetermined position that follows each reserved slot, reinforcing the positional invariance crucial for KV cache reuse.
Second, we apply a Policy-Consistent Attention Masking strategy, as exemplified by the partial mask matrix at the bottom right of Figure~\ref{fig:main}, to accurately simulate the streaming visibility constraints. Similar to SimulMask~\cite{raffel-etal-2024-simultaneous}, during fine-tuning, we modify the attention masks for each target token to mask out source tokens that are not yet available according to the selected simultaneous translation policy (e.g., wait-k or read-n). This forces the model to rely solely on the visible source prefix when predicting the next target token, effectively emulating the incremental information flow of actual simultaneous decoding. By exposing the model to these constraints during training, we ensure that the learned generation behavior is strictly aligned with the streaming dynamics enforced at inference.

\section{Experiments}
\subsection{Setup}
To evaluate the effectiveness of ExPosST, we conduct experiments on five language pairs from the IWSLT 2017 dataset~\cite{cettolo-etal-2017-overview}: English to French (En-Fr), German (En-De), Dutch (En-Nl), Italian (En-It), and Romanian (En-Ro). For all compared methods and baselines, we adopt \texttt{Llama-3.1-8B-Instruct}~\cite{dubey2024llama} and \texttt{Qwen2.5-7B-Instruct} as the backbone LLM. We utilize Low-Rank Adaptation (LoRA)  for parameter-efficient fine-tuning across all experimental configurations. Detailed hyperparameters and LoRA~\cite{DBLP:conf/iclr/HuSWALWWC22} settings are provided in Appendix~\ref{app:hyperparameter}.

We evaluate the effectiveness of ExPosST by comparing it with the following baselines:

\begin{itemize}
\item \textbf{GPE~\cite{tong-etal-2025-llm}:} It adopts the group-streaming paradigm, in which source and target tokens are independently assigned positional encodings to maintain the internal relative order within each sequence. To assess its effectiveness, the performance is evaluated under the \textit{wait-k} policy, with a separate model trained for each $k$ and language pair.
\item \textbf{Conversational SimulMT \cite{wang-etal-2025-conversational}:} This baseline adopts the conversation prompt structure for incremental decoding. The training data is constructed via sentence segmentation using a word alignment tool. During inference, the performance is evaluated under the read-n \& incremental decoding policy \cite{wang-etal-2024-simultaneous}, where a single model is trained for each language pair.
\item \textbf{Offline:} We also report the performance of a non-simultaneous translation model, obtained by performing supervised fine-tuning (SFT) on full sentence pairs and using standard offline translation prompts during inference.
\end{itemize}

To facilitate a direct comparison with the respective baselines, the performance of ExPosST is reported under two configurations: ExPosST(wait-k) and ExPosST(read-n). For ExPosST(wait-k), evaluation is conducted under the \textit{wait-k} policy with $k \in \{1, 3, 5, 7\}$ to align with the GPE setting. In the ExPosST(read-n) setting, we align the Conversational SimulMT settings to build the conversational training data \cite{wang-etal-2025-conversational}. The model is subsequently evaluated following the read-n \& incremental decoding policy \cite{wang-etal-2024-simultaneous}, with $n$ selected from $\{3, 5, 7, 9, 11, 13\}$.

All fine-tuning is implemented within the Simul-LLM~\cite{agostinelli-etal-2024-simul} framework. Inference is carried out using the Simul-LLM \cite{agostinelli-etal-2024-simul} agent integrated with the SimulEval evaluation toolkit \cite{ma-etal-2020-simuleval}.  
We use detokenized BLEU scores computed with SacreBLEU~\cite{post2018call} for the quality metric, and latency is quantified was determined using Length-Adaptive Average Lagging (LAAL)~\cite{papi-etal-2022-generation}.  
\begin{figure}
    \centering
    \includegraphics[width=0.9\linewidth]{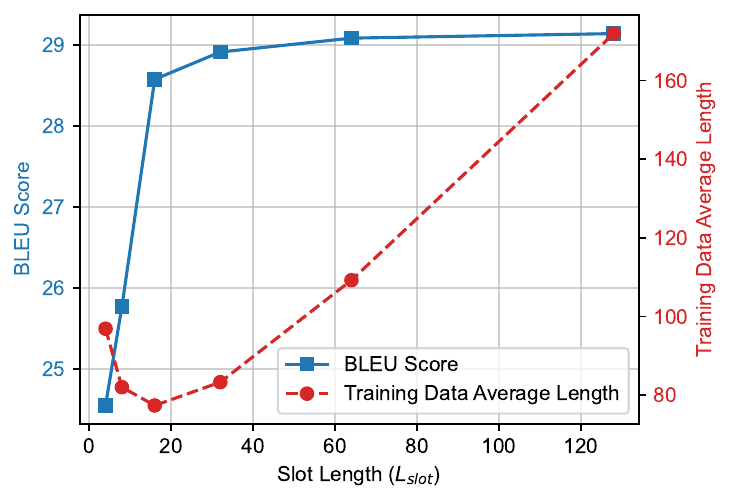}
    \caption{Sensitivity analysis of the pre-allocated slot length $L_{slot}$ on the IWSLT 2017 En-De dev set, illustrating its impact on translation performance (BLEU score) and average training sequence length for the \texttt{Llama-3.1-8B-Instruct} model.}
    \label{fig:slot_size}
\end{figure}

\begin{figure*}[t]
\centering
    \begin{subfigure}{0.32\textwidth}
    \centering
    \includegraphics[width=1\linewidth]{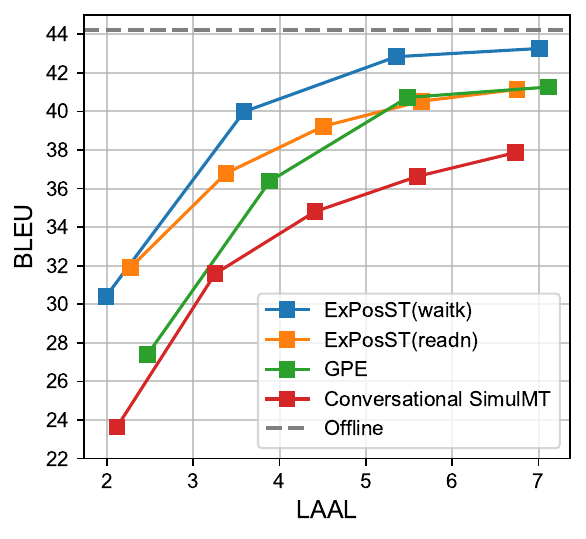}
    \caption{En-Fr}
    \end{subfigure}
    \begin{subfigure}{0.32\textwidth}
    \centering
    \includegraphics[width=1\linewidth]{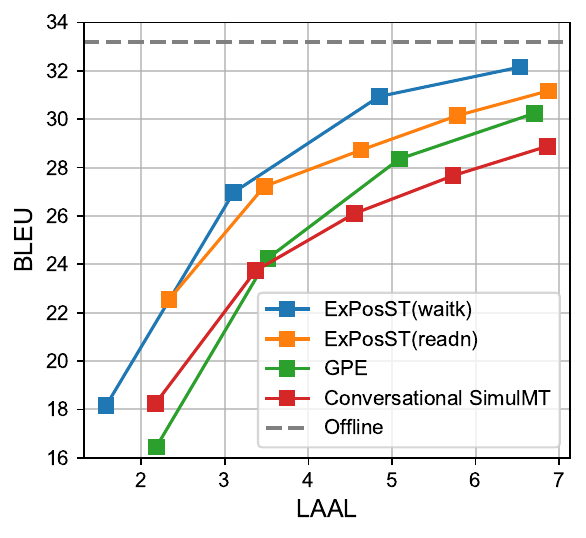}
    \caption{En-De}
    \end{subfigure}
    \begin{subfigure}{0.32\textwidth}
    \centering
    \includegraphics[width=1\linewidth]{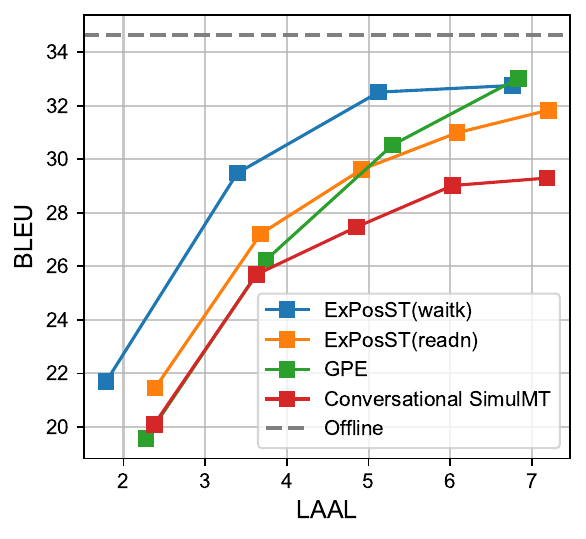}
    \caption{En-Nl}
    \end{subfigure}
    
    \begin{subfigure}{0.32\textwidth}
    \centering
    \includegraphics[width=1\linewidth]{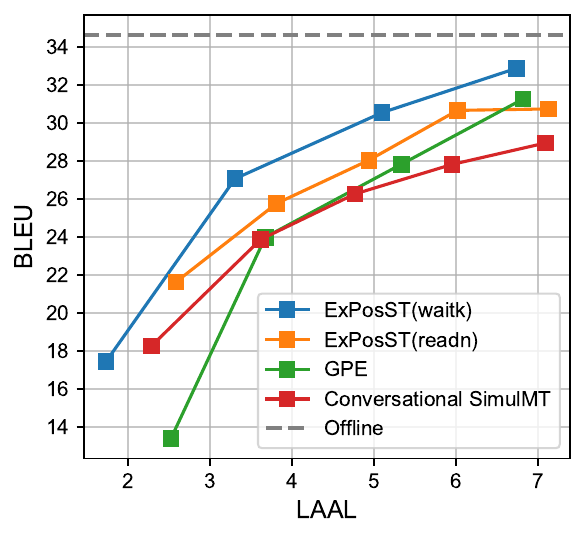}
    \caption{En-It}
    \end{subfigure}
    \begin{subfigure}{0.32\textwidth}
    \centering
    \includegraphics[width=1\linewidth]{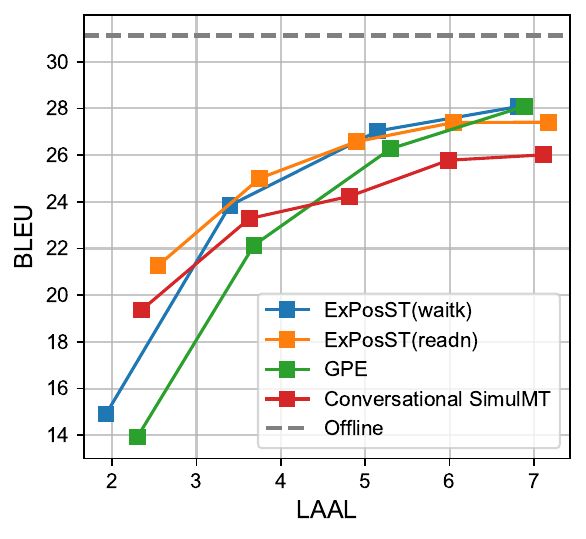}
    \caption{En-Ro}
    \end{subfigure}

\caption{Main results on IWSLT 2017 tasks on \texttt{Llama-3.1-8B-Instruct}. The figures illustrate the BLEU-LAAL trade-off curves, comparing ExPosST with various baselines across two mainstream LLM architectures. The dashed horizontal lines indicate the performance of the corresponding offline models. Higher curves and those shifted toward the top-left represent a superior quality-latency balance.}
\label{fig:mainresults}
\end{figure*}

\subsection{Effect of $L_{slot}$}
\label{sec:lslot_effect}

To determine the optimal configuration for the pre-allocated slot length $L_{slot}$, we conduct a sensitivity analysis on the En-De dev set using the \texttt{Llama-3.1-8B-Instruct} model. We evaluate $L_{slot}$ values across the set $\{4, 8, 16, 32, 64, 128\}$ under wait-7 policies. As illustrated in Figure \ref{fig:slot_size}, the average training sequence length exhibits a characteristic U-shaped trend. Specifically, when $L_{slot}$ is excessively small, the frequent slot allocation and role-switching necessitate a higher density of conversational special tokens (e.g., role markers and delimiters), leading to an increase in total sequence length. Conversely, larger slot lengths require substantial padding to satisfy the fixed capacity of each reserved block, which also escalates the computational footprint. Experimental results demonstrate that translation quality begins to plateau as $L_{slot}$ approaches 16, while the average training sequence length reaches its minimum at approximately $L_{slot}=16$. Meanwhile, the average training sequence length follows a U-shaped trend, reaching its minimum at approximately $L_{slot}=16$. To strike an optimal balance between translation performance and training efficiency, we select $L_{slot}=16$ as our default hyperparameter.

\subsection{Results on Different Language Pairs}

Figure \ref{fig:mainresults} presents the BLEU–LAAL trade-off curves across five language pairs on \texttt{Llama-3.1-8B-Instruct}. 
Under the wait-k policy, ExPosST achieves superior translation quality, with improvements of up to 2–3 BLEU points over the GPE baseline at equivalent latency levels.
Notably, these performance gains exhibit remarkable consistency across all evaluated language pairs. 
As the value of $k$ increases, the performance of ExPosST progressively approaches that of the offline one, which translates the full sentence without streaming constraints. 
In the ``\emph{read‑n \& incremental decoding}'' policy, ExPosST yields a uniform improvement of over 1 BLEU point compared to the conversational SimulMT baseline across the entire latency range.
These results demonstrate that ExPosST is a robust framework, effectively maintaining a superior quality-latency Pareto frontier regardless of the specific decoding policy or target language.

\begin{figure}[t]
\centering
    \begin{subfigure}{0.4\textwidth}
    \centering
    \includegraphics[width=1\linewidth]{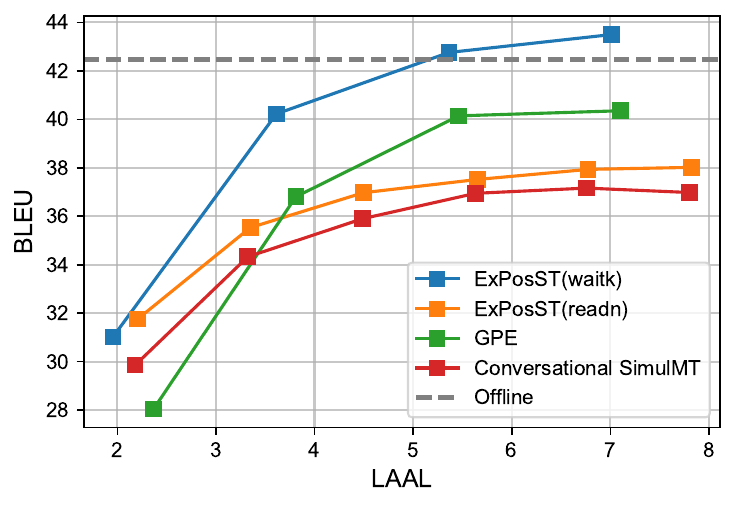}
    \caption{En-Fr}
    \end{subfigure}
    \begin{subfigure}{0.4\textwidth}
    \centering
    \includegraphics[width=1\linewidth]{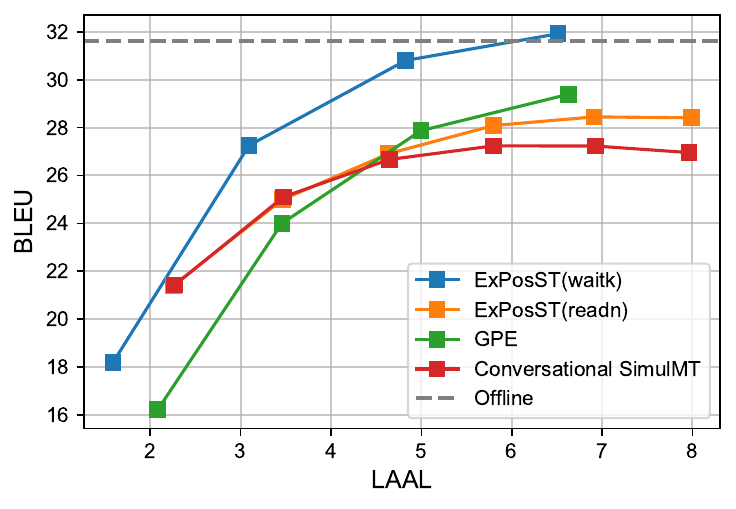}
    \caption{En-De}
    \end{subfigure}

\caption{BLEU-LAAL trade-off on IWSLT 2017 datasets on \texttt{Qwen2.5-7B-Instruct}.}
\label{fig:qwenresults}
\end{figure}
\subsection{Results on Different Models}

To further evaluate the generalization capabilities of the ExPosST framework across diverse architectures, we conducted supplementary experiments using the \texttt{Qwen2.5-7B-Instruct} model. These evaluations were performed on the English-French (En-Fr) and English-German (En-De) translation tasks from the IWSLT 2017 dataset. As illustrated in Figure \ref{fig:qwenresults}, our proposed method consistently achieves performance that is on par with or significantly exceeds the established baselines across the quality-latency spectrum, demonstrating the robust efficacy of ExPosST across a variety of model architectures.

\begin{figure}[t]
    \centering
    \includegraphics[width=0.7\linewidth]{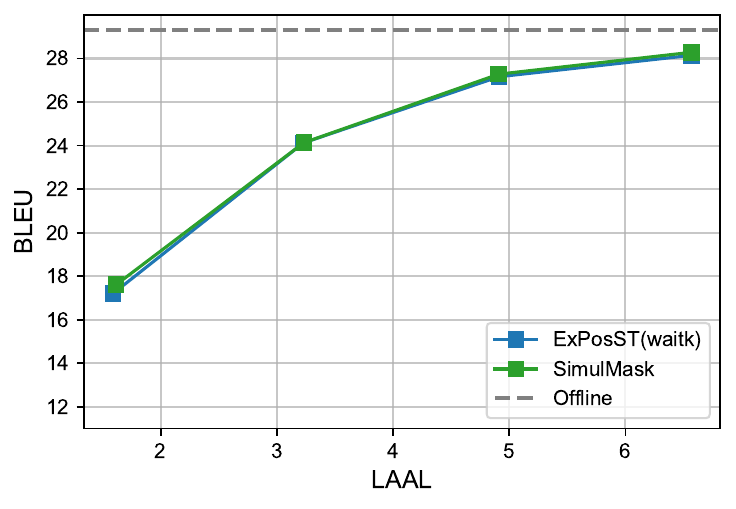}
    \caption{The BLEU vs LAAL result in \texttt{falcon-rw-1b} between SimulMask and ExPosST on En-De.}
    \label{fig:falcon}
\end{figure}

To enable a direct comparison with SimulMask~\cite{raffel-etal-2024-simultaneous}, which is a method specifically designed for ALiBi-based models, we conducted additional experiments on the ALiBi-based \texttt{falcon-rw-1b} model. 
Both SimulMask and our ExPosST framework are evaluated under the wait‑$k$ policy with $k\in$\{1,3,5,7\}, following the $k$ training setting in SimulMask \cite{raffel-etal-2024-simultaneous}.
As shown in Figure \ref{fig:falcon}, ExPosST achieves performance comparable to SimulMask on the \texttt{falcon-rw-1b} model, indicating that our framework adapts effectively to alternative positional encoding schemes without significant degradation.

\section{Analysis}
\begin{figure}[ht]
    \centering
    \includegraphics[width=0.85\linewidth]{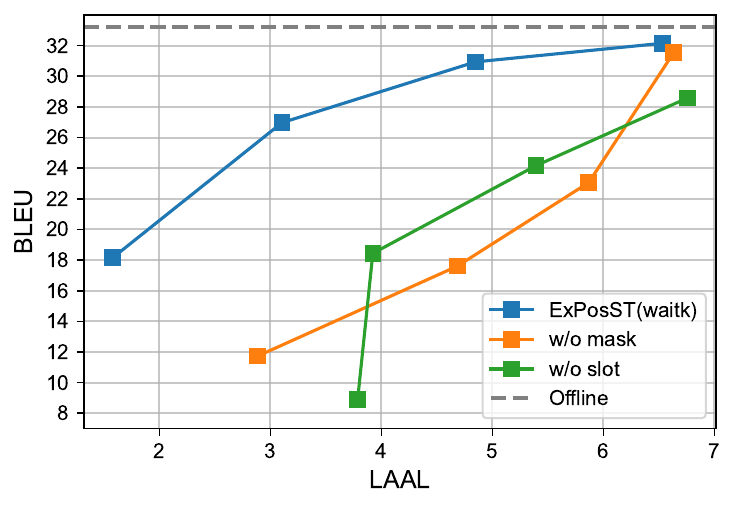}
    \caption{Ablation study of \texttt{ExPosST} components on the En-De task using \texttt{Llama-3.1-8B-Instruct}. The curves represent the quality-latency trade-off for the full model versus variants where masking or slot allocation is removed.}
    \label{fig:ablation}
\end{figure}
\subsection{Ablation Study}
\label{sec:ablation}

To investigate the individual contributions of the core components in the ExPosST framework, we conducted an ablation study using the \texttt{Llama-3.1-8B-Instruct} model on the En-De task. We compared the full ExPosST (wait-$k$) framework against two variants:

\begin{itemize}
    \item \textbf{w/o Masking}: This variant retains the Pre-allocated Positions Inference strategy and the data segmentation into slots during fine-tuning, but removes the policy-consistent attention masking.
    \item \textbf{w/o Slot}: This variant employs the attention masking strategy to simulate simultaneous translation but removes the pre-allocated position slot mechanism. This setup typically leads to positional misalignment during incremental updates.
\end{itemize}

Figure~\ref{fig:ablation} illustrates the BLEU-LAAL trade-off curves for these variants. The results demonstrate that removing either the masking strategy or the position slots leads to a significant degradation in translation quality. Specifically, without masking, the model fails to learn the incremental generation behavior required for low-latency scenarios. In contrast, removing the slot mechanism disrupts positional consistency during inference, confirming that both components are essential to achieve an optimal balance between inference efficiency and translation accuracy.

\subsection{Impact of Training-Inference Slot Mismatch}

\begin{table}[t]
    \centering
    \begin{tabular}{c|cccc}
    \hline
    $L_{slot}$     & wait-1 & wait-3 & wait-5 & wait-7 \\
    \hline
     4 & 12.94 & 19.81 & 13.31 & 13.23 \\
     8 & 16.84 & 25.92 & 29.18 & 26.81 \\
     \textbf{16} & 18.16 & 26.97 & 30.94 & 32.15 \\
     32 & 17.82 & 27.11 & 31.08 & 32.39 \\
     64 & 15.94 & 26.79 & 31.07 & 32.46 \\
     \hline
    \end{tabular}
    \caption{Sensitivity analysis of the mismatch between training and inference slot lengths ($L_{slot}$) on the IWSLT 2017 En-De dataset using \texttt{Llama-3.1-8B-Instruct}. The model is fine-tuned with a fixed training slot length of $L_{slot}=16$ and evaluated under various wait-k policies with different inference slot lengths.}
    \label{tab:mismatch}
\end{table}
We further investigate the sensitivity of ExPosST to discrepancies between the slot lengths used during fine-tuning and inference. This analysis is conducted on the IWSLT 2017 En-De dataset using the \texttt{Llama-3.1-8B-Instruct} model. We fix the training slot length at $L_{slot}=16$ and evaluate the translation performance across a range of inference slot lengths.

As illustrated in Table~\ref{tab:mismatch}, we observe an asymmetric impact on translation quality resulting from configuration mismatches. Specifically, when the inference slot length is significantly smaller than the training setting (e.g., $L_{slot}=4$ or $8$), the BLEU scores suffer a substantial degradation. This suggests that insufficient positional spacing disrupts the attention mechanism's ability to accurately locate the source context segments learned during training.

Interestingly, when the inference slot length exceeds the training value (e.g., $L_{slot}=32$ or $64$), the performance remains relatively stable, with only marginal fluctuations observed across various wait-k policies. This phenomenon indicates that while the model develops a dependency on the positional distributions established during SFT, it exhibits a higher degree of robustness to larger positional offsets than to compressed ones. Nevertheless, since peak performance is consistently achieved when configurations align, maintaining consistency between training and inference remains the optimal strategy for preserving the efficacy of the ExPosST framework.

\subsection{Computational Cost}

\begin{figure}[t]
    \centering
    \includegraphics[width=0.9\linewidth]{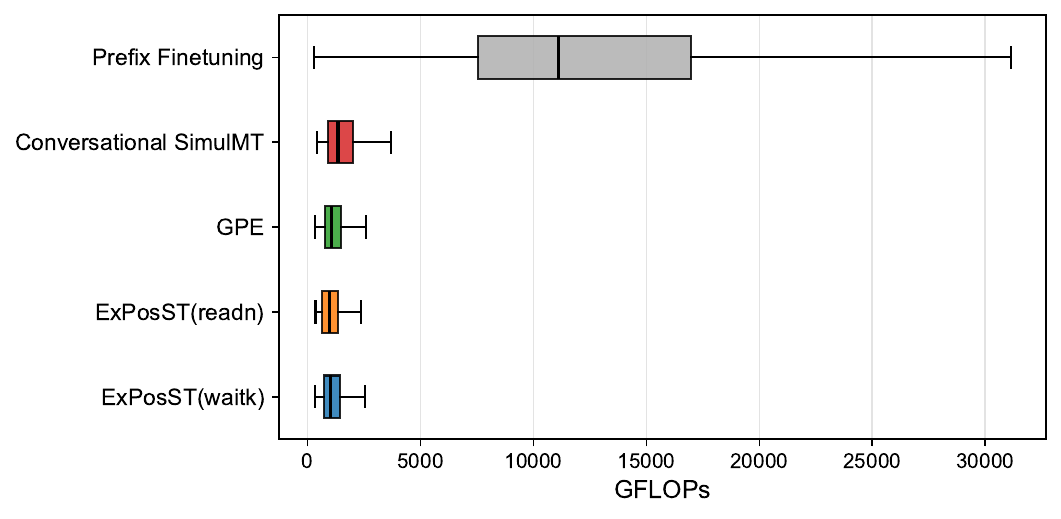}
    \caption{Box plots of cumulative computational cost (GFLOPs) during inference on the IWSLT 2017 En-De test set using \texttt{Llama-3.1-8B-Instruct}. We report results under the wait-3 and read-5 policies.}
    \label{fig:computation}
\end{figure}

We evaluate the computational efficiency of ExPosST during the inference phase. To provide a comprehensive comparison, we measure the cumulative floating-point operations (GFLOPs) using the \texttt{Llama-3.1-8B-Instruct} model on the IWSLT 2017 En-De test set. Specifically, we compare our framework against various baselines under the wait-3 and read-5 policies, incorporating the Prefix Finetuning~\cite{ma-etal-2019-stacl} method as a primary baseline to highlight the overhead of positional re-computation.

As illustrated in Figure~\ref{fig:computation}, there is a distinct hierarchy in computational costs. The Prefix Finetuning method exhibits the highest computational overhead and the largest variance. This inefficiency stems from the necessity of re-encoding the entire sequence at each step to maintain positional consistency, which negates the benefits of KV caching. While the Conversational SimulMT approach permits cache reuse, its reliance on a conversation structure introduces frequent role-switching tags. This leads to "prompt bloating," which increases the sequence length and significantly elevates attention-related computation costs.

In contrast, ExPosST (across both wait-k and read-n configurations) achieves the lowest computational expenditure. By leveraging the pre-allocated position slots, our framework ensures zero-recomputation KV cache reuse while eliminating the overhead associated with redundant conversational markers. Experimental results confirm that ExPosST significantly reduces the median GFLOPs compared to the Conversational baseline. This superior efficiency demonstrates that ExPosST provides a highly hardware-efficient solution for real-time simultaneous translation without compromising quality.

\section{Related Work}
\label{sec:related_work}

Simultaneous Machine Translation (SimulMT) has undergone a significant paradigm shift, transitioning from encoder-decoder architectures~\cite{ma-etal-2019-stacl,ma2019monotonicmultiheadattention,arivazhagan-etal-2019-monotonic,DBLP:conf/iclr/ZhangF23} to decoder-only Large Language Models (LLMs)~\cite{wang-etal-2024-simultaneous,koshkin-etal-2024-llms}. A foundational strategy for SimulMT is prefix fine-tuning~\cite{ma-etal-2019-stacl}, which enables real-time translation by training models to predict target tokens based on partial source prefixes. However, directly adapting this paradigm to decoder-only LLMs~\cite{agostinelli-etal-2024-simul,DBLP:conf/iclr/YuZZXZZ25} introduces a critical architectural challenge: the incremental arrival of streaming source tokens shifts the positional indices of subsequent tokens. This shift causes a structural mismatch with the cached Key-Value (KV) states. This leads to a fundamental dilemma between inference efficiency and positional consistency, prompting recent research to diverge into two primary trajectories.

The first category focuses on adapting internal model mechanisms, such as attention masks and positional encoding schemes~\cite{raffel-etal-2024-simultaneous,ouyang2024fasstfastllmbasedsimultaneous,tong-etal-2025-llm}. For instance, SimulMask~\cite{raffel-etal-2024-simultaneous} introduces simultaneous attention masking during fine-tuning to emulate inference-time behavior. While such an approach successfully enables cache reuse, its reliance on ALiBi positional encodings~\cite{press2022trainshorttestlong} restricts portability to mainstream LLMs that utilize Rotary Positional Embeddings (RoPE)~\cite{SU2024127063}, such as the Llama and Qwen series. Similarly, Group Position Encoding~\cite{tong-etal-2025-llm} assigns independent positional ID groups to source and target segments to avoid positional misalignment. However, the relative positional intervals between source tokens and target tokens in GPE may be zero or fractional, which deviate from standard LLM pre-training objectives.

The second category reformulates sequence layouts by employing conversation prompt to append tokens incrementally~\cite{wang-etal-2025-conversational,ouyang-etal-2025-infinisst,fu2025llmsachievehighqualitysimultaneous,fu2025efficientadaptivesimultaneousspeech}. Conversational SimulMT~\cite{wang-etal-2025-conversational} utilizes a framework where source and target slots interleave as conversations, ensuring that all incoming tokens are strictly appended to the end of the existing sequence, thereby facilitating KV cache reuse. Building on this concept, EAST~\cite{fu2025llmsachievehighqualitysimultaneous} incorporates explicit read-write signals to switch between operations adaptively. Despite their efficiency, the frequent role-switching inherent in these methods can distract the model from the source context. Furthermore, EAST necessitates high-quality synthetic data to construct its specialized training sets.

In contrast to these existing approaches, we introduce ExPosST, a framework that ensures both inference efficiency and positional consistency through explicit position allocation. By pre-allocating fixed ``positional slots'' for incoming source tokens, ExPosST maintains invariant positional indices for the target sequence during inference. This design enables zero-recomputation KV cache reuse while remaining fully compatible with mainstream RoPE-based architectures, eliminating the need for unnatural conversation prompt formats or specialized positioning modifications.

\section{Conclusion}
In this work, we introduce ExPosST, a general framework designed for adapting decoder-only LLMs to SimulMT. The core of ExPosST is an Explicit Position Allocation strategy that pre-allocates fixed ``\emph{position slots}'' for streaming source tokens, thereby ensuring that the positional indices of the target sequence remain invariant during READ/WRITE cycles. Our design fundamentally resolves the positional dilemma in KV cache reuse, enabling zero-recomputation inference while maintaining strict positional consistency. Furthermore, we introduce a Policy-Consistent Fine-tuning strategy that aligns training supervision with the slot-based streaming protocol and simulates the incremental visibility constraints of simultaneous policies, ensuring that the model learns to translate effectively under realistic latency conditions. Experimental results and in-depth analysis strongly demonstrate the effectiveness and generality of our framework.


\bibliographystyle{named}
\bibliography{ijcai26}

\appendix

\section*{Appendix}

\section{Hyperparameters}
\label{app:hyperparameter}

The fine-tuning hyparameters of each baseline is shown in Table \ref{tab:hyperparameter}
\begin{table}[h!]
    \centering
    \begin{tabular}{ccc}
    \hline
     \textbf{Hyperparameter} & \textbf{Group 1} \\ 
     \hline
     \textbf{Weight Precision} & bfloat16   \\
    \textbf{Optimizer} & AdamW \\
    \textbf{Learning Rate} & $2\cdot10^{-4}$ \\
    \textbf{LR Scheduler} & Inverse Sqrt \\
    \textbf{Weight Decay} & 0.1  \\
    \textbf{Warmup Steps} & 500   \\
    \textbf{Max Gradient Norm} & 1  \\
    \textbf{Max Sequence Length} & 512 \\
    \textbf{Epochs} & 2   \\
    \textbf{Batch size} & 64   \\
    \textbf{Lora Rank} & 32 \\
    \textbf{Lora Alpha} & 16 \\

    \hline
    \end{tabular}
    \caption{Fine-tuning hyperparameters for all models.}
    \label{tab:hyperparameter}
\end{table}

\begin{figure*}[!h]
\centering
    \begin{subfigure}{0.32\textwidth}
    \centering
    \includegraphics[width=1\linewidth]{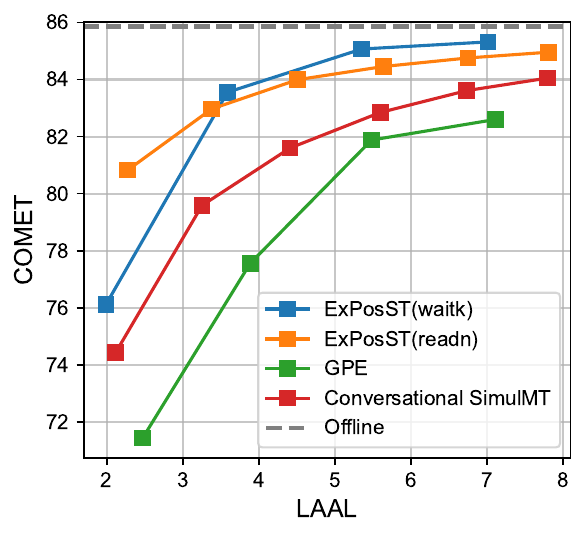}
    \caption{En-Fr}
    \end{subfigure} 
    \begin{subfigure}{0.32\textwidth}
    \centering
    \includegraphics[width=1\linewidth]{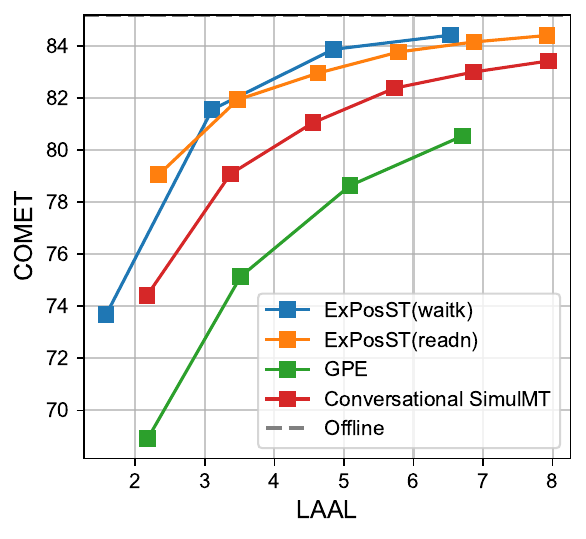}
    \caption{En-De}
    \end{subfigure}
    \begin{subfigure}{0.32\textwidth}
    \centering
    \includegraphics[width=1\linewidth]{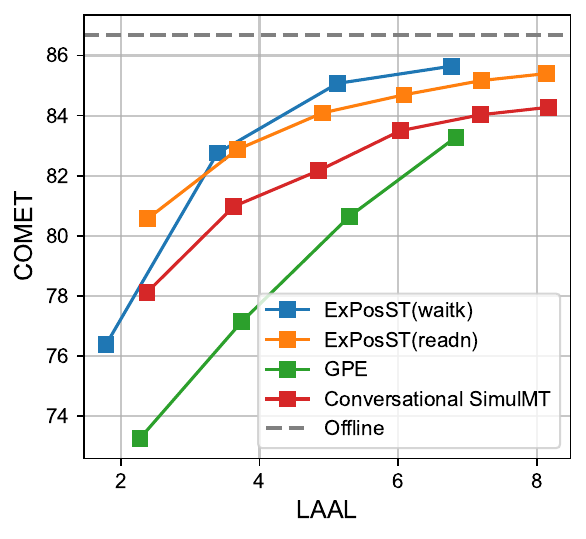}
    \caption{En-Nl}
    \end{subfigure}
    
    \begin{subfigure}{0.32\textwidth}
    \centering
    \includegraphics[width=1\linewidth]{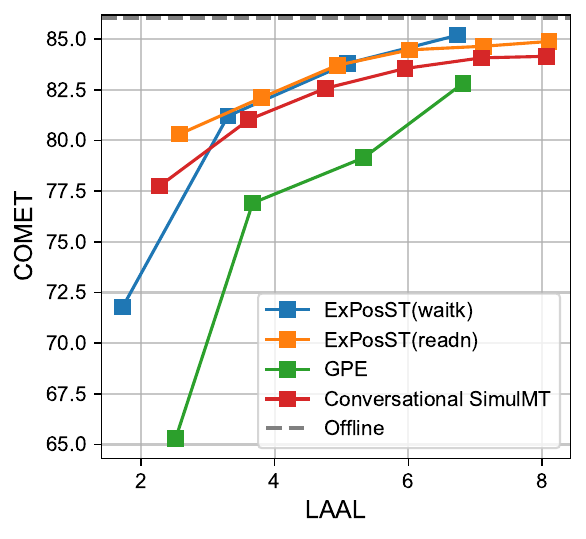}
    \caption{En-It}
    \end{subfigure}
    \begin{subfigure}{0.32\textwidth}
    \centering
    \includegraphics[width=1\linewidth]{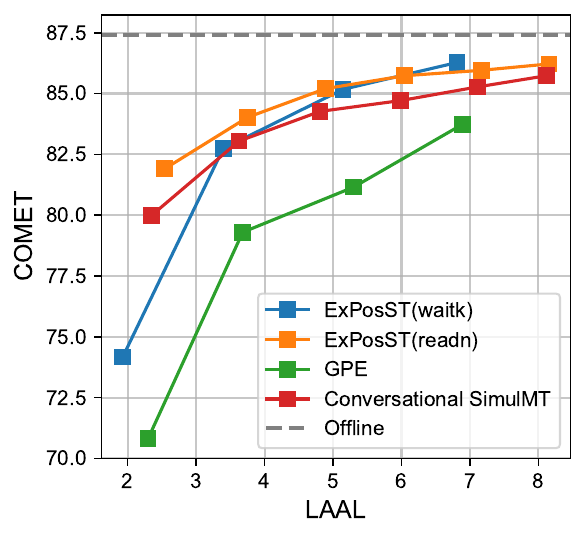}
    \caption{En-Ro}
    \end{subfigure}

\caption{COMET results on IWSLT 2017 tasks on \texttt{Llama-3.1-8B-Instruct}. The curves demonstrate the trade-off between translation quality (COMET) and latency (LAAL) for ExPosST against various baselines.}
\label{fig:comet}
\end{figure*}
\section{Results on COMET scores}
\label{sec:appendix_comet}

To provide a more comprehensive evaluation of translation quality beyond n-gram overlap, we report the results using the COMET metric on the IWSLT 2017 tasks. As shown in Figure~\ref{fig:comet}, ExPosST consistently outperforms all baseline methods across various languages and latency constraints.

Specifically, both ExPosST-waitk and ExPosST-readn establish a superior quality-latency Pareto frontier. Even at low-latency regimes (low LAAL), ExPosST maintains high semantic fidelity, significantly narrowing the gap between simultaneous and offline translation. These results reinforce the conclusion that explicit position allocation preserves the model's instruction-following and translation capabilities more effectively than existing internal modification or conversational prompting strategies.

\end{document}